# Markovian Reeb Graphs for Simulating Spatiotemporal Patterns of Life

Anantajit Subrahmanya[1], Chandrakanth Gudavalli[1], Connor Levenson[1], and B.S. Manjunath[1]

[1] University of California, Santa Barbara, Santa Barbara, CA, USA

**Abstract.** Accurately modeling human mobility is critical for urban planning, epidemiology, and traffic management. In this work, we introduce Markovian Reeb Graphs, a novel framework that transforms Reeb graphs from a descriptive analysis tool into a generative model for spatiotemporal trajectories. By embedding probabilistic transitions within the Reeb graph structure, our approach captures individual and population-level Patterns of Life (PoLs) and generates realistic trajectories that preserve baseline behaviors while incorporating stochastic variability. We present two variants —Sequential Reeb Graphs (SRGs) for individual agents and Hybrid Reeb Graphs (HRGs) that combine individual with population PoLs, which we evaluate on the Urban Anomalies and Geolife datasets using five mobility statistics. Results demonstrate that HRGs achieve strong fidelity across metrics while requiring only modest trajectory datasets without specialized side information. This work establishes Markovian Reeb Graphs as a scalable, data-efficient framework for trajectory simulation with broad applicability across urban environments.

## 1 Introduction

Modeling and synthesizing human mobility patterns is essential for urban planning [1, 2], traffic management [3], energy allocation [4], public health [5], and disaster preparedness [5]. While the availability of mobile devices and location-based services has enabled large-scale GPS data collection [6, 7], privacy restrictions and limited longitudinal datasets hinder robust analysis and generalization. Even the largest publicly available datasets [8, 9] capture only a narrow slice of behavior, motivating the need for simulation frameworks that can extrapolate from limited observations. Traditional simulation methods, such as Activity-Based Models (ABMs) [10] encode activity schedules and travel demand but require extensive calibration and high computational overhead, limiting scalability and adaptability. Deep learning approaches [10, 11] address some challenges by learning population-level mobility patterns from large datasets, but they remain tied to specialized, high-volume data sources (e.g., financial transactions, social media check-ins), restricting their generalizability across urban environments and ability to generate trajectories for individual agents pertaining to their Pattern of Life (PoL). Recently, Reeb graphs have emerged as a promising method for analyzing geospatial trajectories [12]. Prior work has shown that Temporal Reeb Graphs can partition trajectories into meaningful substructures for anomaly detection [13], suggesting that Reeb graphs encode Patterns of Life (PoLs). However, existing Reeb graph formulations are primarily descriptive: they detect deviations but cannot generate realistic trajectories, nor do they differentiate frequent from rare events.

Our key contribution is to transform Reeb graphs from an analysis tool into a generative framework. We introduce Markovian Reeb Graphs, which embed probabilistic transitions within the Reeb graph structure to model individual- and population-level mobility. This enables the generation of realistic future trajectories that preserve baseline PoLs while incorporating stochastic variability. Unlike ABMs, our method doesn't require extensive scenario-specific calibration, and unlike deep learning approaches, it can operate directly on modest trajectory datasets without specialized side information. This is the first work to unify topological representations of mobility with probabilistic modeling for scalable trajectory simulation.

We now summarize the main contributions of the paper:

1. We introduce Markovian Reeb Graphs, a data structure capturing individual and population-level Patterns of Life (PoLs).
2. We present a novel and performant simulation method using Markovian Reeb Graphs.
3. We evaluate generated trajectories on agent-level and population-level metrics both qualitatively and quantitatively.

## 2 Related Works

### 2.1 Past work on Reeb Graphs

Reeb graphs are structures which track changes in level set topology over a scalar function for shape analysis [14]. Recent scholarship adapted Reeb graphs for topological modeling of trajectories, including the analysis of neuronal fibers [15, 16] and human mobility patterns [12, 13], to summarize dense sequences as a sparse graph.

Previous work on geospatial trajectories has shown that Temporal Reeb Graphs (TERGs) can partition GPS points into nodes and edges to represent invariant subtrajectories for anomaly detection [13]. TERGs are constructed over time by defining nodes at points of change in the connected trajectories — when two or more trajectories transition from spatially disconnected to connected (a "connect" event) or vice versa (a "disconnect" event). Under this model, trajectories that disconnect from a Reeb graph are considered out of distribution, hence anomalous. The ability of TERGs to identify out of distribution points implies that Reeb graphs capture PoLs and could be used for simulating additional trajectories conforming to existing patterns of life for an individual agent.

Exploring Reeb graphs' utility beyond individual agent PoLs has been explored Multi-Agent Reeb Graphs (MARGs), which capture population-level behaviors from multiple agents' trajectories. Like the single agent case, trajectories disconnecting from the MARG are considered anomalous, implying a MARG could simulate PoLs at a population level.

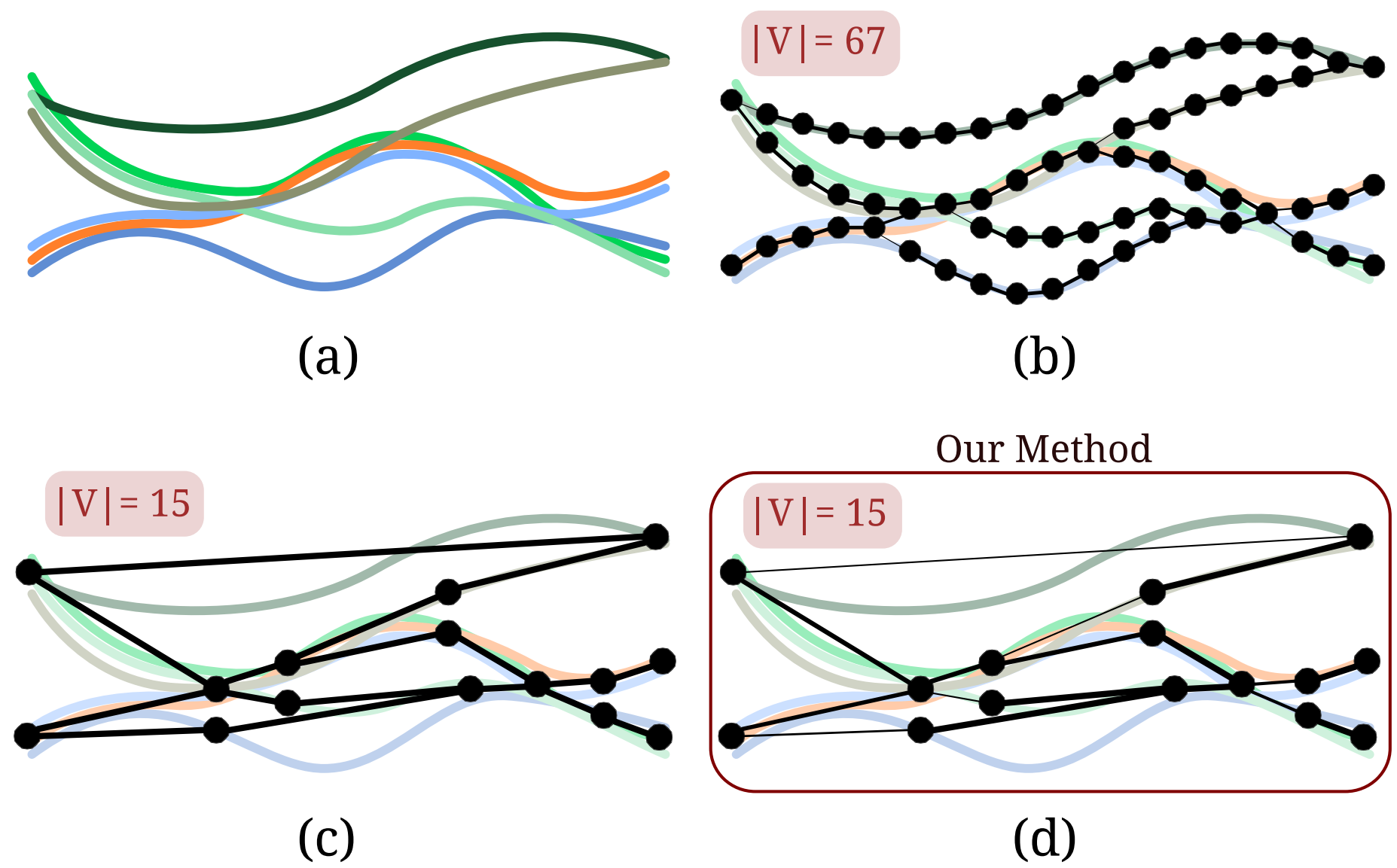

**Fig. 1.** (A) Seven 1D trajectories (vertical axis: position, horizontal axis: time). (B) Time-Dependent Markov Chain constructed on trajectories with edge thickness proportional to weight. This captures path probabilities at cost of complexity (high node/edge counts) (C) TEmporal Reeb Graph (TERG) constructed on trajectories. This has low complexity but cannot differentiate between common/rare events. (D) Markovian Reeb Graphs (our method) capture both the variance and frequency of the trajectories. Edge thickness is proportional to weight.

While MARGs and TERGs capture the possibility of an event occurring through connect and disconnect events, they cannot differentiate between frequent and infrequent events (Fig. 1C). Therefore, using existing Reeb graph data structures for any human mobility simulation would confound rare and daily behaviors, leading to unrealistic trajectories. Inspired by work modeling the human connectome [16], Markovian Reeb Graphs resolve this limitation by including probabilities of a deviation in behavior through graph edge weights (Fig. 1D).

### 2.2 Past Work on Mobility Simulation Engines

Routing engines are used to simulate the physical movement of agents between start and end points. Some allow setting waypoints and selecting the mode of transport to coarsely control the agent's velocity. For example, Valhalla [17] is an open-source routing engine for multi-modal trip planning, producing realistic point-to-point routes on real-world road networks. Although Valhalla computes plausible paths given an origin, destination, and travel mode, it does not model temporal activity patterns, agent-level variability, or higher-level behavioral rules. Thus, Valhalla-generated trajectories are limited to shortest-path or fastest-path behaviors and lack the diversity and stochasticity of real human mobility. The Simulation of Urban MObility (SUMO) [18] mitigates some of these limitations by offering a higher resolution modeling framework with increased degrees of freedom. SUMO supports detailed control over traffic lights, vehicle interactions, and mobility demand modeling. However, realistic SUMO simulations require extensive input data such as accurate (traffic) demand matrices, calibrated departure times, and high-fidelity behavioral parameters. Furthermore, calibrating and scaling SUMO to large agent populations can be computationally intensive. Unlike Valhalla and SUMO, which focus on route planning and traffic simulation and require extensive prior information, our Markovian Reeb Graph model learns individual and population mobility dynamics from observed trajectories and efficiently extrapolates this information to generate additional trajectories.

Many mobility engines use Activity-Based Models (ABMs) to model complex daily schedules for synthetic populations [19, 20]. ABMs are used in active commuting research, generating daily plans for agents (i.e. location labels at each time). MATSim uses a co-evolutionary algorithm where agents optimize their daily plans (changing route, times, modes of transport, etc.) to maximize a "utility score" until reaching equilibrium. The Patterns of Life Simulation [21] uses Maslow's hierarchy of needs [22] to influence agents' decisions and interactions based on psychological, safety and social needs. Despite optimization efforts [19], ABMs remain a computationally intensive simulation method, with super-linear scaling by the number of agents. Emerging deep generative models, like DeepAM [10], use machine learning to generate trip patterns. DeepAM is trained on regional population information, household travel surveys and data from existing ABMs. However, unlike ABMs, the inference cost for generating each trajectory is constant regardless of the number of agents. Unlike both types of simulation engines, Markovian Reeb Graph models don't require specialized data sources and can extrapolate from a small number of simulated or real trajectories.

Scaling limitations in collecting large volumes of specialized data have motivated recent works to explore learning complex urban mobility patterns from GPS trajectories [23–25]. Deep generative models like MoveSim [26] and DeepMobility [27] utilize trajectory data and semantic information to learn population dynamics and generate new agent trajectories, evaluated using group-level Jensen-Shannon Divergence for statistics like distance traveled per day, radius of gyration, and travel duration. Markovian Reeb Graphs excel on these metrics and can simulate future movements of existing agents.

Many of these methods compare performance against the long-withstanding Markov chain approach [28]. Mobility Markov Chain (MMC) models [28–30], where POI locations represent states, are extremely limited because they do not capture any route or temporal information [31]. Weighted Markov prediction models have chosen an alternative state assignment method, capturing coarse route information by assigning the nearest cell tower as the state [30]. However, such methods don't capture any temporal information either. One proposed solution is through Time Varying Markov Chains (TVMCs), where the transition probabilities between different states varies based on the time of day [32]. These are topologically equivalent via a lifting transformation to traditional Markov chains with a unique state for each time point [33]. Notably, TVMCs require side information (agent classifications, occupation, etc) to form the time varying components – however, given a sufficiently high spatial and temporal resolution, TVMCs could theoretically capture movement between stop points and route information taking time of day into account. However, such models would be impractical due to having an extremely high number of states (Fig. 1B). Markovian Reeb Graphs capture equivalent information to TVMCs with an order of magnitude fewer states: this offers a scalable method of trajectory generation conforming to agent PoLs, accounting for route preferences and time-of-day-dependent behaviors.

## 3 Methods

We provide detailed descriptions of the algorithms to generate trajectories conforming to agent-level and population-level distributions using three types of Markovian Reeb Graphs. First, we introduce Sequential Reeb Graphs (SRGs), an improvement on TERGs which capture the frequency of behaviors in an individual agent's PoL. Next, we introduce a hybrid approach using Multi-Agent Reeb Graphs (MARGs) to generalize SRGs to represent population-level behaviors. Finally, we explain how Markovian Reeb Graphs are used to generate trajectories.

### 3.1 Sequential Reeb Graph Construction (SRG)

A trajectory $T$ is a sequence of points (tuples) $p_0, p_1, ..., p_L$ with length $L$. Each $p_i \in T$ consists of (index, latitude, longitude) triplets; these may be replaced with semantic embeddings that better represent agent and population patterns of life [34]. Each agent has $N$ trajectories.

Two points $p_i = T_1[i]$ and $q_j = T_2[j]$ belong to a "bundle" (equivalence class) if $d(p_i, q_j) < \varepsilon$ and $i = j$ for some metric $d$ and threshold $\varepsilon$. For this paper, we assume $d$ is the Euclidean distance between the latitude/longitude pairs. We define the set of bundles $B = \{b_1, b_2, ..., b_m\}$. By our formulation above, all points in a bundle $(p_a, p_b, ...) \in b_i$ correspond to the same index across their source trajectories – let this quantity be defined as $\text{index}(b_i)$. Each bundle has a well-defined centroid, computed by averaging the components of the points within, defined as $\text{centroid}(b_i)$. Finally, we define $\text{cc}(b_i) = \{k : T_k[\text{index}(b_i)] \in b_i\}$, the set of "connected" trajectories containing values in $b_i$

The set of nodes in an SRG is defined by changes in the sets of connected trajectories over the indices of the underlying trajectories. Each node $v \in V \subset B$ must satisfy $\text{cc}(v) \neq \text{cc}(b)$ for any $b \in B$ such that $\text{index}(b) = \text{index}(v) - 1$ – equivalently, a bundle at index $i$ is also a node if there's no bundle in the previous time step with the same connected components. Each directed edge in the Reeb graph from $v_i$ to $v_j$ satisfies $\text{cc}(v_i) \cap \text{cc}(v_j) \neq \emptyset$ with edge weight $w = \frac{\#(\text{cc}(v_i) \cap \text{cc}(v_j))}{\#(\text{cc}(v_i))}$ where $\#$ is the number of elements in each set. Intuitively, the edge weight represents the conditional probability that a trajectory is in bundle $v_j$ given it was in bundle $v_i$, making this Reeb graph "probabilistic" in nature.

Therefore, any SRG can be characterized by a set of $N$ trajectories $A = \{T_1, T_2, ..., T_N\}$ and two hyper-parameters: $d$ (a distance metric) and $\varepsilon$ (a scalar threshold). Inspired by past works [13] on Reeb graphs for spatiotemporal data, our algorithm for constructing SRGs has two phases:

1. Bundle Computation: For all points across trajectories, compute a partition $B$ such that each $b \in B$ forms a bundle defined by $d$ and $\varepsilon$.
2. Graph Generation: Compute the subset $V \subset B$ and the set of edges $(v_i, v_j, w) \in E$.

For computing bundles, we used an incremental algorithm [13] for bundle computation using R-trees (a spatial data structure) to store the bundles themselves, with time complexity $O(LN \log(N))$.

### 3.2 Population-Individual Hybrid Reeb Graphs

It is possible to construct a Markovian Multi-Agent Reeb Graph (MARG) that captures population-level patterns using $K$ agents $\{A_1, A_2, ..., A_k\}$ with $N$ trajectories each by concatenating the agents to construct a set of $KN$ trajectories, each with length $L$: $M = \{T_1, T_2, ..., T_{KN}\}$. Hence, the time complexity to compute a MARG is $O(LKN \log(KN))$. As with the SRG, the expected complexity scales log-linearly with the number of bundles; hence MARGs are faster to compute for groups of similar agents.

To encode both population and individual PoLs in a Reeb graph, we merge the MARG and SRG for each agent to create a Hybrid Reeb Graph, capturing an individual agent's past PoL and realistic deviations from that PoL using population data.

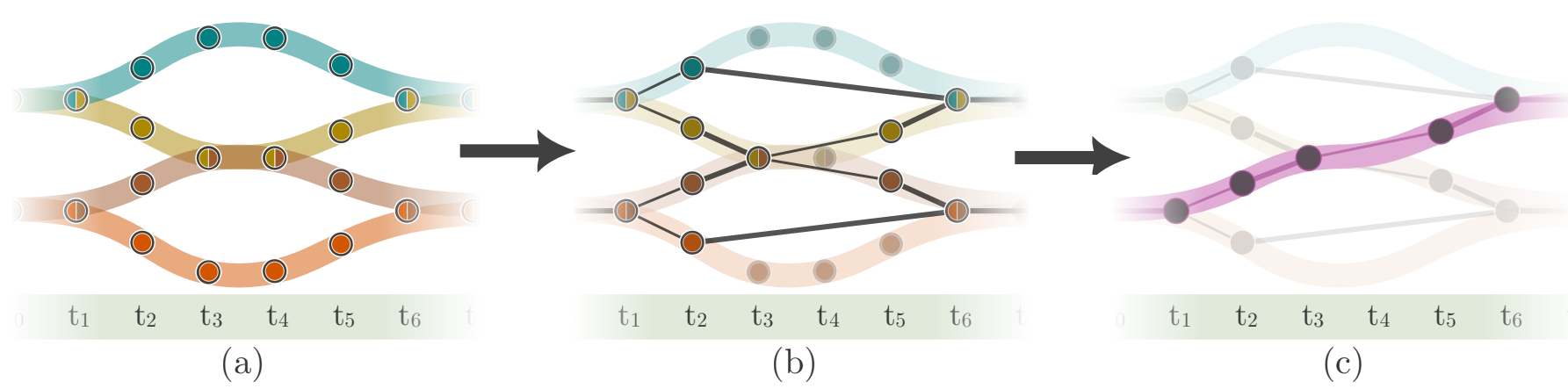

**Fig. 2.** Process of generating trajectories with Markovian Reeb Graphs. (a) Visualization of bundles (circles) for a section of four trajectories. Two trajectories are "connected" if they overlap. The colors shown within each bundle represent which trajectories belong to the bundle at each time step. (b) Sequential Reeb Graph computed from the previously computed bundles. Edge thickness corresponds to the edge weights (in this example, either $\frac{1}{2}$ or 1). (c) A trajectory generated by the Sequential Reeb Graph. We generate a random traversal of the Reeb Graph's nodes and edges, then copy a random subtrajectory corresponding to two edges. Concatenating all such subtrajectories results in a new trajectory (pink).

Let $S(V_s, E_s)$ be the SRG of agent $A_k$, and let $M(V_m, E_m)$ be the population's MARG. We will fuse these graphs to create an HRG $H(V_h, E_h)$ for this agent. First, we find the corresponding nodes and edges between $S$ and $M$. That is, for each node in $V_s$, we find the node with the nearest centroid in $V_m$ – note that the nodes cannot differ by more than $\varepsilon$ since $A_k$ is represented in $M$. We draw an edge with weight $\alpha$ from each node in the SRG to its corresponding node in the MARG, and an edge with weight $\beta$ from each node in the MARG to its corresponding node in the SRG. Intuitively, $\alpha$ represents the probability of an agent deviating from their PoL, and $\beta$ represents the probability of an agent reconnecting to their original Reeb graph. To anchor the HRG to $A_k$'s PoL, any critical point in $H$ must have a path back to the agent SRG and any graph traversal must begin on the SRG for continuity between trajectories. Only the nodes and edges that satisfy both requirements were retained from $M$ to $H$.

A critical limitation of a naive merge of an individual and population level graph is that if a traversal of the hybrid graph leaves the SRG, it may not reconnect with the SRG for many samples. This would lead to trajectories gradually drifting further and further from an agent's PoL over time. To account for this, we adjusted the weight of each edge in the HRG proportional to the expected time to return to the SRG. After all weight adjustments, the graph was traversed once to ensure the sum of outbound edge weights for every node is 1. Thus, the complexity of constructing the HRG is linear with respect to the nodes and edges in the MARG.

### 3.3 Trajectory Generation with Markovian Reeb Graphs

Sequential Reeb Graphs, Multi-Agent Reeb Graphs, and Hybrid Reeb Graphs can generate trajectories through random node traversal in the Reeb Graph. Each edge $(u, v)$ between two nodes of the Markovian Reeb Graph corresponds to a set of subtrajectories transitioning between critical points, $\mathrm{cc}(u) \cap \mathrm{cc}(v)$. We generate each subtrajectory by randomly picking one corresponding to the edge. This traversal continues, concatenating these subtrajectories until ultimately a disappear/sink node (with no outbound edges) is reached.

To generate a new trajectory using any Markovian Reeb Graph, we start with an empty trajectory $T = []$. We select the nearest source node to the endpoint of the previous trajectory to ensure consistency between trajectories (i.e., consecutive days). Next, we randomly select one of the outgoing edges of the chosen node, with the selection probability equal to its weight. We find the set of trajectories defining each node, say $\mathrm{cc}(u)$ and $\mathrm{cc}(v)$, and select one at random, say $T_i$. We copy the values of $T_i$ corresponding to this edge, the subtrajectory $T_i[\mathrm{index}(u) : \mathrm{index}(v)]$, and append this to $T$. We continue this process, picking an edge from $v$ until $v$ is a sink node, at which point $T$ should have a length of $L$. The time complexity of generating each trajectory will be linear with the maximum path length of the graph.

## 4 Experiments

### 4.1 Datasets

We evaluate our methods on two publicly available large-scale mobility datasets. For each dataset, we consider two subsets of training (M1) and provided evaluation (M2) subsets, alongside the data generated by our method (G).

- Geolife [9]: The Geolife dataset contains real data from 182 users over a period of five years collected by the MSRA Geolife project (April 2007 to August 2012). It contains 17,621 GPS trajectories near Beijing, China, with varying trajectory counts and durations for each user.
- Urban Anomalies (UA) [22]: Each UA subset is a synthetic dataset containing one month of normal data (M1) and another month of potentially anomalous data (M2) in either Atlanta or Berlin. We discarded agents with an anomaly in M2, leaving 28 days of data for 880 agents with a GPS sampled every 5 minutes (total of 7M points) for each subset.

Sequential Reeb Graphs require uniformly sampled data, so we re-sampled the Geolife dataset to generate a sample every 10 seconds using linear interpolation (8640 samples per trajectory). If there were no samples within 1 minute of a time point, we considered it a dropout, represented by setting the latitude and longitude to infinity. This allowed us to construct Reeb Graphs while accounting for dropout. We then applied a post-processing step to remove all nodes at infinity and their corresponding edges to prevent dropout from affecting the Reeb graph trajectories. This technique allows us to adapt Sequential Reeb Graphs to sparsely populated data. M1 and M2 were defined manually, such that an equal number of trajectories were present in each subset.

### 4.2 Evaluation Metrics

Following the common practice in previous works [11, 31, 35], we define 6 statistics to quantitatively evaluate our method's performance: **Daily Travel Distance:** the cumulative travel distance, **Daily Travel Duration:** the number of moving samples, **Shortest Distance Traveled:** the shortest trip distance (or 0 if no trip is present), **Longest Distance Traveled:** the longest trip distance (or 0 if no trip is present), **Radius of Gyration:** spatial range of movement and **New Locations Visited:** the number of unique stop point locations in M2/G which were not present in M1.

We evaluated these statistics at the population and individual levels. The agent-level metrics were computed for M1, M2, and the Reeb graph trajectories G. We then computed the agent-wise mean and the Mean Absolute Error between M1 and M2, and M1 and G. The population-level metrics were generated in a similarly, except we computed a histogram with 100 bins on each statistic across all agents, then computed the Jensen-Shannon Divergence between the distributions of M1 and M2, and M1 and G. This measured how well the generated trajectories adhere to each agent's PoL, as well as how well they conform to population Patterns of Life.

Due to the high rate of dropout for the Geolife dataset, we computed distances on M1 and M2 on a subset of the time range when agents had the most data points (roughly 5.5 hours) and kept agents with less than 15 consecutive minutes of dropout during this period in M1, M2 and G (n = 24). We filled in the missing timestamps with the previous value, quantized at one sample every 10 seconds.

It is important to recognize that lower MAE/JSD implies stronger conformance with M1, however, this is not necessarily desirable because PoLs naturally evolve over time. For this reason, the desired result should be for our MAE/JSD measures to be *near* those of M2, as marked in Table 1 and Table 2.

### 4.3 Results

Table 2 and Table 1 include computed values for all six metrics using an SRG ($\varepsilon = 10^{-3}$) and an HRG ($\varepsilon = 10^{-3}, \alpha = 1\%, \beta = 99.9\%$). These hyperparameter values were chosen empirically. An ablation study is included in Supplementary Materials. Due to limited per-agent data in the Geolife dataset, we were only able to generate full trajectories for a subset of agents. We considered the M2 score as a baseline and marked the generation method (SRG, HRG) closer to the baseline M2 value. For Geolife, the HRG outperformed the SRG on all metrics. This may be due to the SRG lacking information to travel to all locations visited due to high dropout in M1, while the HRG can utilize MARG information to connect multiple disconnected locations. In contrast, the Urban Anomalies dataset performance was mixed. Both SRG and HRG performed similarly for distance traveled, shortest and longest trip distances, and radius of gyration. However, the HRG performed better in movement duration and new locations visited. Note that the latter is unsurprising since SRGs cannot visit new locations not in M1.

**Table 1.** Geolife

| Evaluation Statistic | Generation Method | Agent Score | Population Score |
|---|---|---|---|
| Distance Traveled | M2 | 0.04053 | 0.03816 |
| | HRG | **0.04569** | **0.05129** |
| | SRG | 0.04897 | **0.05129** |
| Duration of Movement | M2 | 35.62602 | 0.10754 |
| | HRG | **47.64929** | **0.15249** |
| | SRG | 64.36842 | 0.19429 |
| Longest Trip Distance | M2 | 16.10039 | 0.34872 |
| | HRG | **16.48089** | **0.36374** |
| | SRG | 16.59337 | **0.36374** |
| New Locations Visited | M2 | 1.03834 | 0.62800 |
| | HRG | **0.67179** | **0.26403** |
| | SRG | 0.00000 | 0.00000 |
| Radius of Gyration | M2 | 0.00669 | 0.03147 |
| | HRG | **0.00791** | **0.04865** |
| | SRG | 0.00831 | **0.04865** |
| Shortest Trip Distance | M2 | 0.00000 | 0.00000 |
| | HRG | **0.00230** | **0.12109** |
| | SRG | 0.00231 | **0.12109** |

Our generated trajectories show strong conformance to population-level average traffic patterns given sufficient data. Observe the frequency of visiting each grid location between M1, M2, and G for the Urban Anomalies dataset are nearly identical. In contrast, the M1, M2, and G heat maps for the Geolife dataset look very different, with M1 as a superset of M2 and M2 as a superset of G. One limitation of our generation method is it constructs continuous trajectories for an entire day, while the original datasets have a large amount of dropout. This means certain locations may be unreachable even with the MARG, if there is no series of trajectories to reach them. This is also observed in the M1 heat map, where distinct “islands” of points appear to be unconnected, meaning an agent starting in one isolated region has no knowledge of a path to another region.

**Table 2.** Urban Anomalies

| Evaluation Statistic | Generation Method | Agent Score | Population Score |
|---|---|---|---|
| Distance Traveled | M2 | 0.00972 | 0.03891 |
| | HRG | **0.00406** | 0.20935 |
| | SRG | 0.00401 | **0.20451** |
| Duration of Movement | M2 | 1.53693 | 0.04087 |
| | HRG | **1.16636** | 0.28211 |
| | SRG | 0.65946 | **0.27314** |
| Longest Trip Distance | M2 | 0.04276 | 0.16495 |
| | HRG | 0.09806 | 0.42455 |
| | SRG | **0.09750** | **0.42416** |
| New Locations Visited | M2 | 0.00081 | 0.01677 |
| | HRG | **0.00057** | **0.01403** |
| | SRG | 0.00000 | 0.00000 |
| Radius of Gyration | M2 | 0.00034 | 0.02781 |
| | HRG | **0.00073** | 0.17383 |
| | SRG | 0.00073 | **0.17381** |
| Shortest Trip Distance | M2 | 0.00538 | 0.09241 |
| | HRG | 0.01529 | 0.41480 |
| | SRG | **0.01497** | **0.39098** |

To understand Reeb Graphs intuitively, we can examine the density of nodes over time in the MARGs (each node has a latitude, longitude, and time point). For both datasets, we computed a histogram of the number of nodes for each time slice (indirectly a measure of variance). We applied a Gaussian kernel to smooth the histograms to understand how the density of the Reeb graphs varied over time. For the Urban Anomalies dataset, there are no nodes between 2am and 6am. As agents "wake up" around 6am, the number of nodes ramps up. There are spikes at 8am (going to work), 1pm (lunch), 5pm (leaving work), and 6pm (dinner) — however, the number of nodes during throughout the movement period is consistently high (Fig. 3A). In the Geolife dataset, agents have greater variations. The Reeb graphs were computed in UTC, but the histogram has been re-aligned to local time UTC+8. The initial spike at 8am is an artifact of Reeb graph construction since a node is always created if it is present at the start and end of the trajectories. We observe a spike near 9am, presumably agents going to work, with little variation throughout the day (agents do not leave work). Near the end of the day, there is an increase in variance as agents leave work, followed by decaying node counts as they return home (Fig. 3B). These trends in the Geolife dataset may be influenced by dropout (nodes are placed at the start and end of dropout periods), but the similar trends in the simulated dataset (no dropout) suggest they may be present despite bias not because of it.

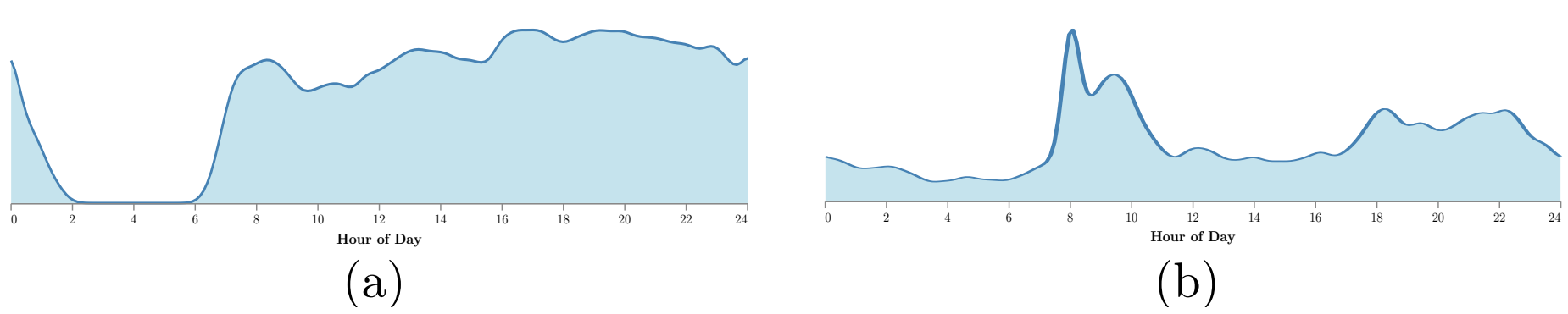

**Fig. 3.** Reeb Graph Statistics. Histograms show the density of nodes in the Urban Anomalies (A) and Geolife (B) dataset over hours of the day.

Strong performance in generating trajectories conforming to population-level PoLs indicates that the MARG of a population may capture sufficient information on the region's dynamics. As a preliminary exploration, we queried the MARG built on the Geolife dataset to find a path between two points and asked for the most probable traversal between those paths within a ten-hour period. We applied our previous trajectory generation algorithm to convert this into a continuous trajectory. The resulting trajectory doesn't exist in the dataset but approximates low-level behaviors like slowing down at intersections (Fig. 4B), navigating one-ways (Fig. 4C), and using multiple modes of transport (Fig. 4A). Although promising, this is a preliminary exploration in the routing engine space and further research is needed before making any authoritative judgments.

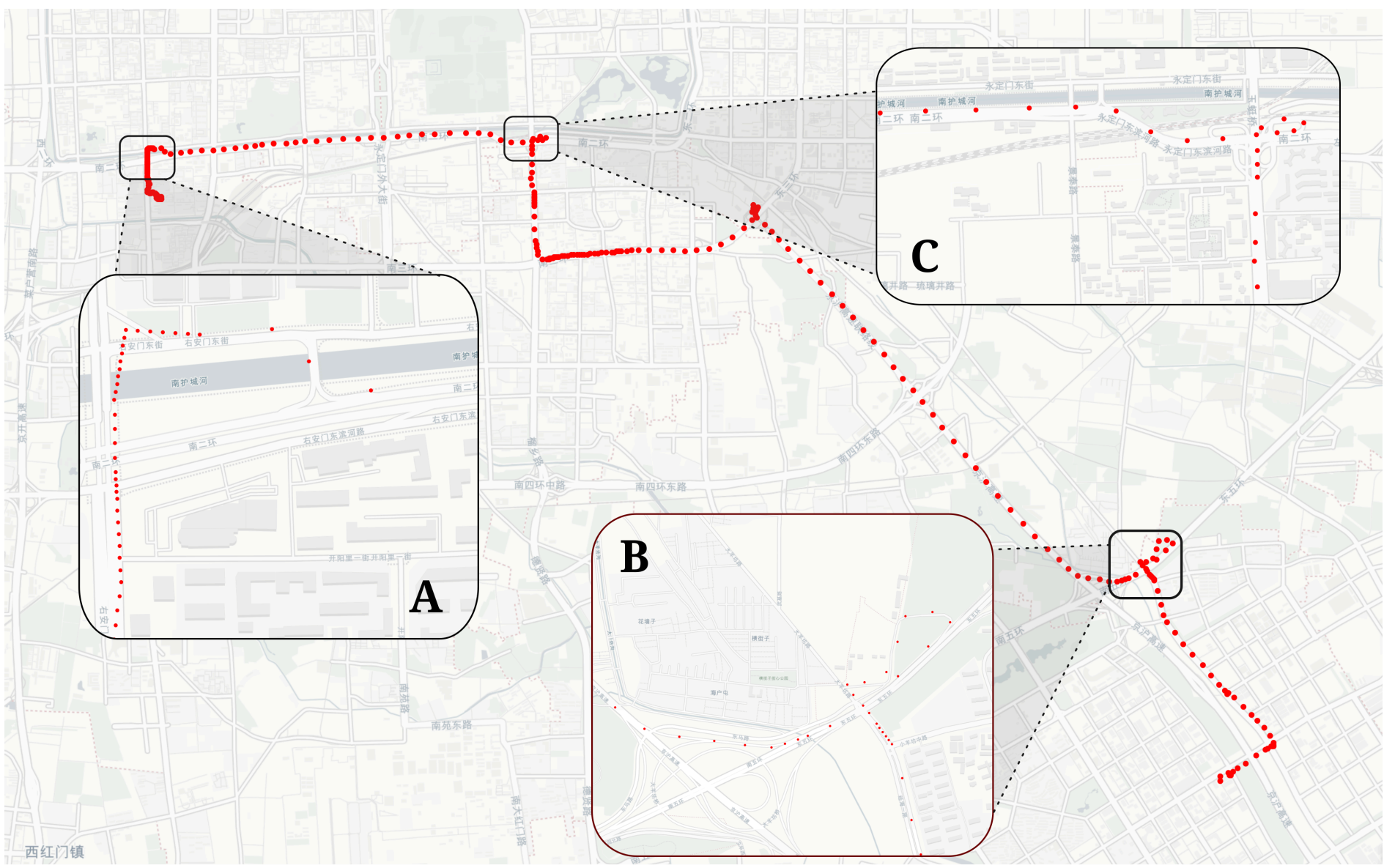

**Fig. 4.** An example trajectory generated using the MARG-based routing method. The algorithm was given two coordinates, a start time (9am), and an end time (10:30am). (A) Agent switches mode of transit, perhaps dismounting from a bus at the bus stop. (B) Agent slows down before crossing a busy intersection. (C) Agent correctly navigates one-way signs and a roundabout.

### 4.4 Reproducibility Statement

All methods to reproduce every figure, table, and result in this paper are open source and well documented at https://github.com/anantajit/MarkovReebs and the two datasets used are publicly available [9, 22]. Each subset (M1, M2, G) is stored as lists of tensors, so researchers can benchmark our generated trajectories using other metrics, and our evaluation metrics can run on trajectories in this representation generated by other methods. We created a standardized interface for interacting with each dataset, so additional datasets can be added to extend this work.

## 5 Conclusion

In this paper, we introduced Markovian Reeb Graphs as a generative framework for modeling human mobility. Through Sequential, Multi-Agent, and Hybrid variants, we demonstrated how Reeb graphs can simulate realistic trajectories that preserve individual and population-level Patterns of Life. Our evaluation on the Urban Anomalies and Geolife datasets showed that Hybrid Reeb Graphs achieve the best balance across agent- and population-level metrics, performing exceptionally well on real world data.

Looking ahead, several extensions remain open. Improving low-level agent conformance, incorporating robustness to noisy and sparse GPS data, and enabling richer semantic embeddings could further enhance realism. Extending the framework to handle periodic routines and assign likelihoods for anomaly detection are also natural next steps. Further exploration into using the MARG to construct temporally-relevant paths between points could reveal additional applications of Markovian Reeb Graphs for complex simulation and out-of-distribution behavior injections. These directions highlight the potential of Reeb-graph–based generative models as a versatile foundation for future mobility synthesis.

## 6 Acknowledgements

This work is supported by the Intelligence Advanced Research Projects Activity (IARPA) via Department of Interior/ Interior Business Center (DOI/IBC) contract number 140D0423C0057. The U.S. Government is authorized to reproduce and distribute reprints for Governmental purposes notwithstanding any copyright annotation thereon. Disclaimer: The views and conclusions contained herein are those of the authors and should not be interpreted as necessarily representing the official policies or endorsements, either expressed or implied, of IARPA, DOI/ IBC, or the U.S. Government. We would like to thank Kin Gwn Lore for insights and assistance during the initial phase of this project.